\documentclass{article}
\usepackage{spconf,amsmath,graphicx}

\usepackage{booktabs}
\usepackage{amsmath}
\usepackage{flexisym}
\usepackage{dcolumn}
\usepackage{kotex}
\usepackage{color}
\usepackage{hhline}
\usepackage{verbatim}
\usepackage{multirow}
\usepackage{amssymb}
\usepackage{rotating}
\usepackage{subfigure}
\usepackage{dblfloatfix}
\usepackage{enumitem}
\newcolumntype{L}[1]{>{\raggedright\let\newline\\\arraybackslash\hspace{0pt}}m{#1}}
\newcolumntype{C}[1]{>{\centering\let\newline\\\arraybackslash\hspace{0pt}}m{#1}}
\newcolumntype{R}[1]{>{\raggedleft\let\newline\\\arraybackslash\hspace{0pt}}m{#1}}

\title{Attentive Modality Hopping Mechanism for Speech Emotion Recognition}
\name{Seunghyun Yoon$^{1}$, Subhadeep Dey$^{2}$, Hwanhee Lee$^{1}$ and Kyomin Jung$^{1}$}
\address{
$^{1}$Dept. of Electrical and Computer Engineering, Seoul National University, Seoul, Korea \\
$^{2}$Idiap Research Institute, Martigny, Switzerland \\
{
\{mysmilesh,~wanted1007,~kjung\}@snu.ac.kr,~subhadeep.dey@idiap.ch
}
}

\begin{document}
\ninept
\maketitle
\begin{abstract}
In this work, we explore the impact of visual modality in addition to speech and text for improving the accuracy of the emotion detection system. The traditional approaches tackle this task by independently fusing the knowledge from the various modalities for performing emotion classification.
In contrast to these approaches, we tackle the problem by introducing an attention mechanism to combine the information. In this regard, we first apply a neural network to obtain hidden representations of the modalities.
Then, the attention mechanism is defined to select and aggregate important parts of the video data by conditioning on the audio and text data.
Furthermore, the attention mechanism is again applied to attend the essential parts of speech and textual data by considering other modalities.
Experiments are performed on the standard IEMOCAP dataset using all three modalities (audio, text, and video). The achieved results show a significant improvement of 3.65\% in terms of weighted accuracy compared to the baseline system.



\end{abstract}
\section{Introduction}
\label{sec:intro}

 
Emotion identification plays a vital role in all human communications.
Our response depends on emotional states, and it also provides nonverbal cues while conveying the message.
Recently, there have been many efforts to understand emotions in computer-assisted technologies~\cite{picard2003affective}.
Automated customer service systems are typically designed to classify the emotions of the speaker to enhance the user experience.
Emotion classification is also beneficial in the paralinguistic field.
Recently, commercial digital assistant applications, such as Siri, found that paralinguistic information, such as emotion, is beneficial for recognizing the intent of the speaker~\cite{mitra2019leveraging}. Humans usually employ multimodality information to identify emotions. However, the impact of multimodality for this area has not yet been fully investigated.
In this paper, we are interested in employing information from textual, speech, and visual models for recognizing emotions.

In the past, different approaches for emotion identification have been explored from speech signals~\cite{han2014speech,lee2011emotion}.
Most speech emotion techniques have focused on extracting low- or high-level features.
In particular, signal processing techniques are applied to extract features, such as cepstral and prosodic features. Suprasegmental features (such as cepstral or prosodic contours) have been shown to provide good performance for this task~\cite{dai2015emotion}. Furthermore, statistical modeling techniques, such as the hidden Markov model (HMM) and the Gaussian mixture model (GMM), have been successfully used for this task~\cite{schuller2003hidden,el2007speech}.

Recently, researchers have explored the application of textual information in addition to speech signals for emotion classification. Lexical information is typically used to search for keywords that express the emotional state of the speaker. In~\cite{gamage2017salience}, lexical information is used by using a bag-of-words representation of the message.
Recent approaches exploit the powerful modeling capabilities of the deep neural network (DNN) for fusing information from both modalities~\cite{yoon2018multimodal,cho2018deep,xu2019learning}. The hidden representations of the modalities are used to combine the knowledge from acoustic and textual data for emotion classification.
In our previous work~\cite{yoon2019speech}, we explored an attention mechanism for exploiting textual information.
The attention mechanism is trained to summarize the lexical content and speech utterance automatically.
Experimental evaluation indicates superior performance on a standard IEMOCAP dataset~\cite{busso2008iemocap}.
In this paper, we extend this approach by incorporating visual information into the framework and proposing an attention mechanism to exploit multimodal knowledge effectively. This is motivated by the fact that humans express emotion through facial expressions, speech signals, and lexical content. We hypothesize that exploiting visual knowledge in addition to speech and lexical information will result in superior performance.

As opposed to combining information from modalities independently, we propose to apply an attention mechanism that aggregates knowledge from one modality conditioned on the other two modalities.
The proposed approach first obtains the sequence of hidden representations from the three modalities (speech, text, and visual).
Then, the summary vector of the visual model is obtained by linearly combining the attention weights with the hidden units.
Furthermore, this vector is then applied to obtain the attention weights of the acoustic data.
The updated visual and acoustic data are consecutively used to compute the attention weight of the textual data to aggregate salient parts of the text modality.
As this process continues multiple times, we hypothesize that the mechanism will effectively compute the relevant parts of each modality.

To evaluate the performance of the proposed approach, we performed emotion recognition experiments on the standard IEMOCAP dataset.
Experiments on this corpus indicate that the proposed approach outperforms the baseline system by 3.65\% relative improvement in terms of weighted accuracy.
In addition, we obtain improved model performance by increasing the iterations over the modality.
The experimental results demonstrate that our proposed model correctly learns to aggregate the necessary information among the modalities via the iterative hopping process.


\section{Recent works}
\label{sec:related}

Recently, several neural network approaches have been successfully applied for emotion classification.
Researchers have proposed convolutional neural network (CNN)-based models that are trained on speech utterances for performing identification~\cite{badshah2017speech, aldeneh2017using}.
There have been some successful approaches using attention mechanisms as well~\cite{li2018attention,mirsamadi2017automatic}. 
In particular, the work in~\cite{mirsamadi2017automatic} presents an approach to incorporate attention-based modeling in the recurrent neural network (RNN) architecture. The attention mechanism is designed to compute the weights or relevance of each frame. An utterance level representation is obtained by temporal aggregation of these weighted speech features. The attention unit is designed to derive the segments of speech for emotion recognition automatically.

Emotion recognition using acoustic and lexical knowledge has also been explored in the literature.
These works have been inspired by the fact that emotional dialogue is composed of not only speech but also textual content. 
 In~\cite{schuller2004speech}, emotional keywords are exploited to effectively identify the classes.
Recently in~\cite{yoon2018multimodal,cho2018deep,sebastian2019fusion}, a long short-term memory (LSTM) based network has been explored to encode the information of both modalities.
Furthermore, there have been some attempts to fuse the modalities using the inter-attention mechanism~\cite{xu2019learning,yoon2019speech}.
However, these approaches are designed only to consider the interaction between acoustic and textual information.




\begin{figure*}[!t]
\small
\centering
\subfigure[AMH-1]{\label{fig_hop_1}\includegraphics[width=0.6\columnwidth]{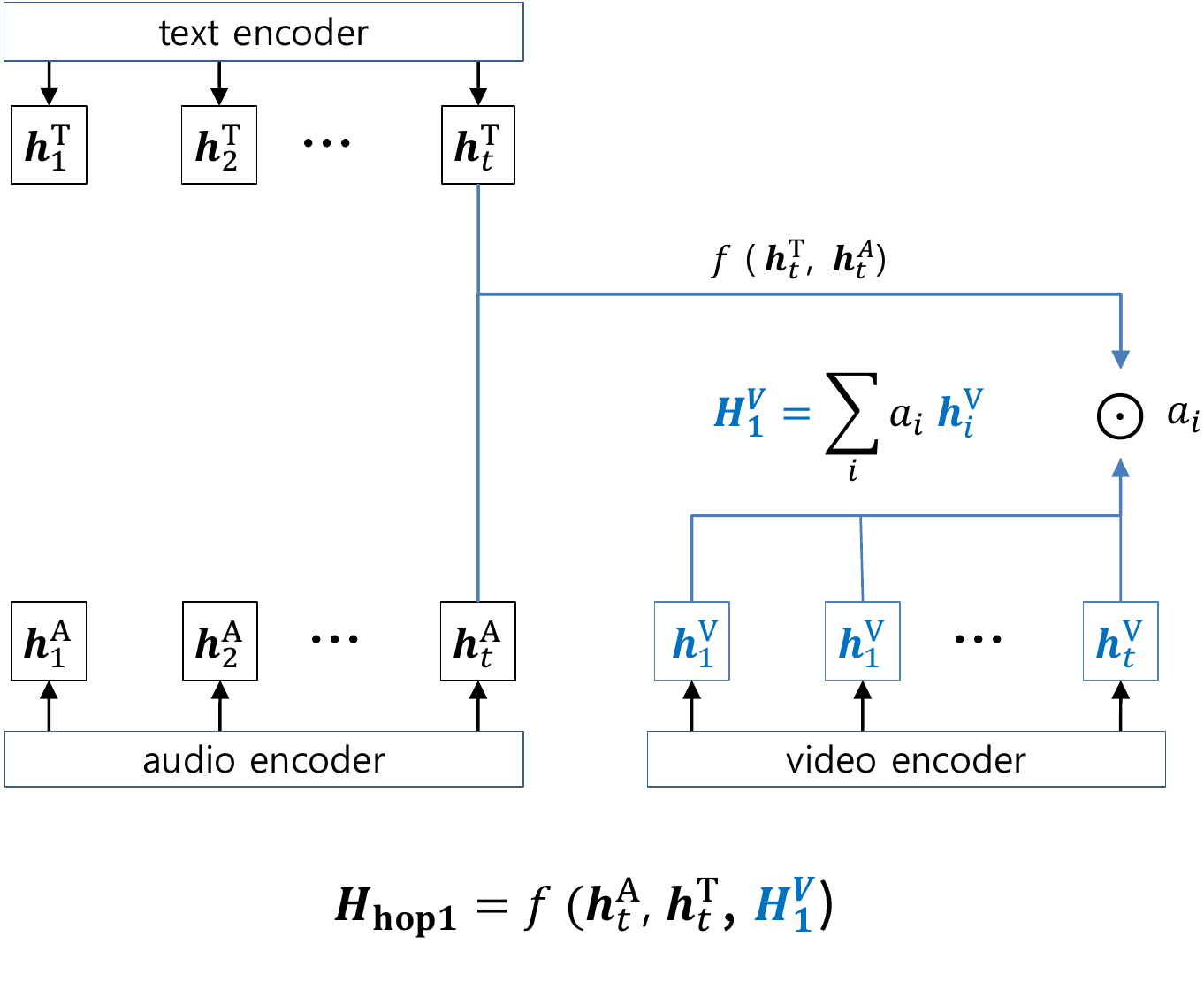}}
\quad
\qquad
\subfigure[AMH-2]{\label{fig_hop_2}\includegraphics[width=0.6\columnwidth]{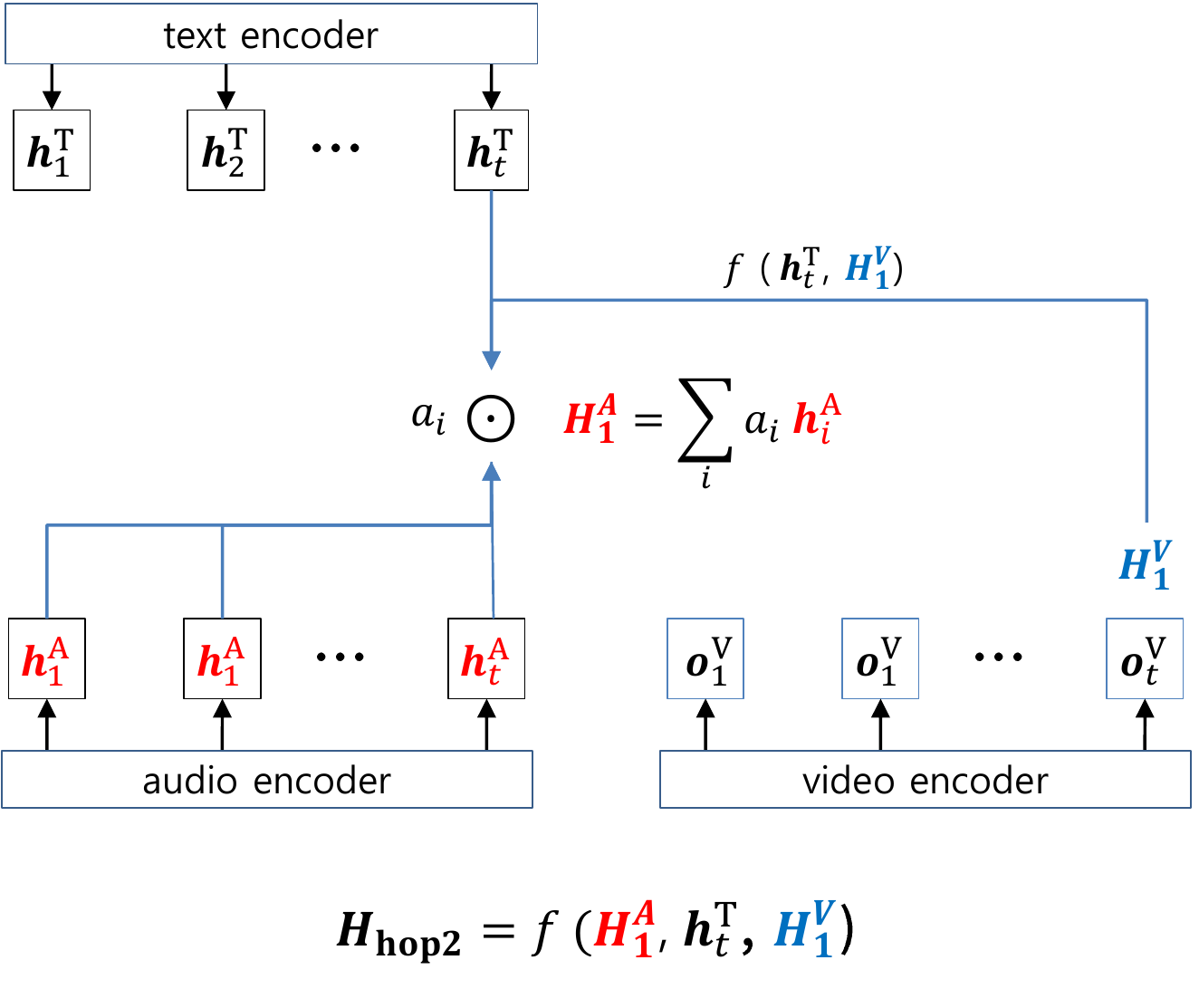}}
\quad
\qquad
\subfigure[AMH-3]{\label{fig_hop_3}\includegraphics[width=0.6\columnwidth]{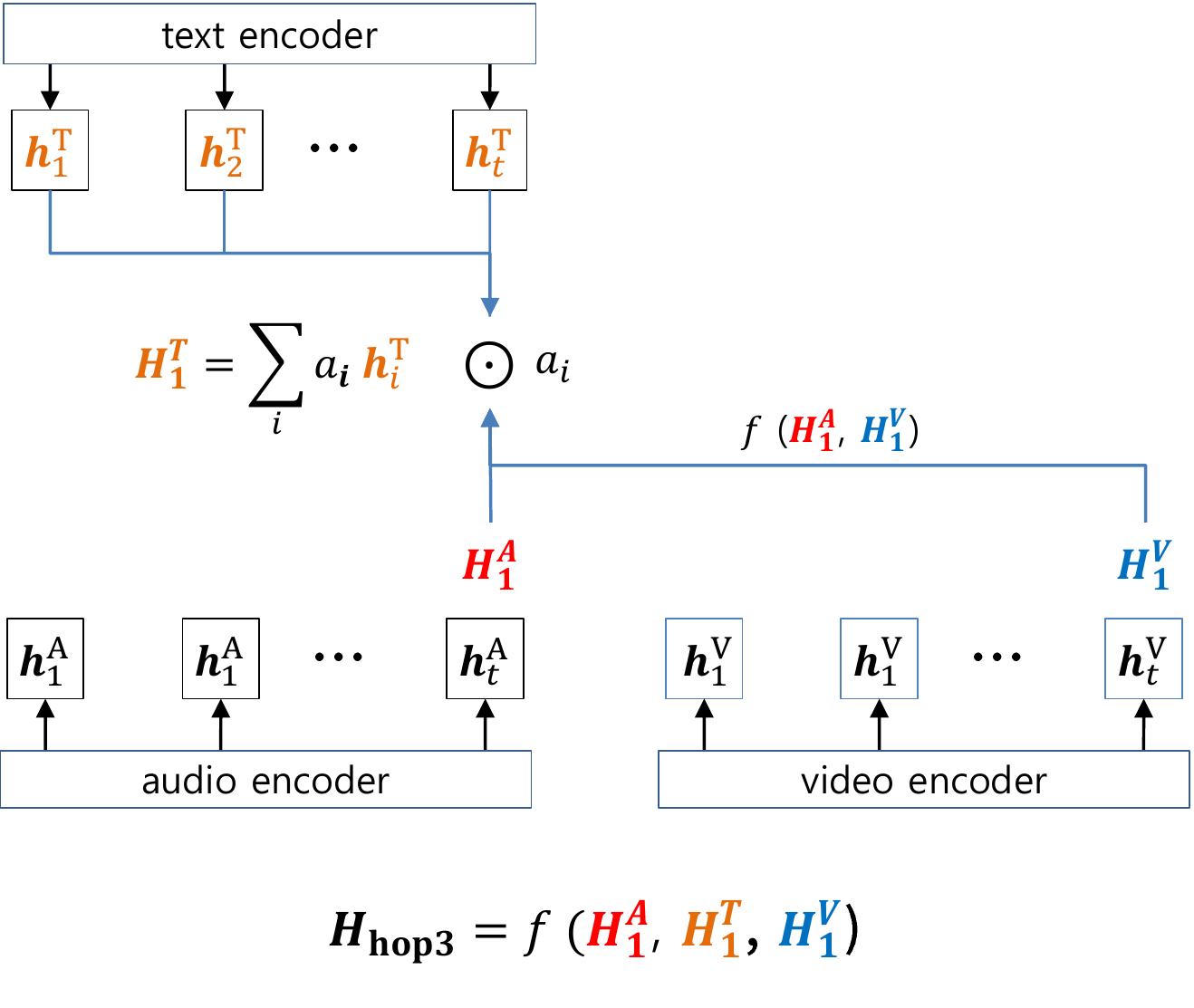}}
\caption{
Architecture of attentive modality-hop mechanism
}
\label{fig_AMH}
\end{figure*}

\section{Model}
\label{sec:model}
This section describes the methodologies that are applied to the speech emotion recognition task.
We start by introducing a recurrent encoder to individually encode the audio, text, and video modalities.
We then propose an approach to exploit each modality one by one.
In this technique, an attentive modality hopping process is proposed to obtain relevant parts of each modality via the iterative aggregation process.

\subsection{Recurrent Encoder}

Motivated by the architecture used in~\cite{yoon2018multimodal,mirsamadi2017automatic, wang2016audio}, we employ a recurrent neural network to encode a series of features in the speech signal and to classify the signal into one of the emotion classes. In particular, we employ a gated recurrent unit (GRU)~\cite{cho2014learning} for each modality (i.e., acoustic, textual, visual) to encode the information, as shown in figure~\ref{fig_AMH}.
The GRU encodes the sequence of the feature vector of each modality by updating its hidden states as follows:

\begin{equation}
\begin{aligned}
& \textbf{h}_{t} = f_{\theta}( \textbf{h}_{t-1}, \textbf{x}_{t}), \\
\end{aligned}
\label{eq_RE}
\end{equation}
where $f_{\theta}$ is the GRU network with weight parameter $\theta$, $\textbf{h}_t$ represents the hidden state at the \textit{t}-th time step, and $\textbf{x}_t$ represents the \textit{t}-th sequential features in a target modality.
This recurrent encoder is used in the same manner as the independent audio, text, and video modalities.
For the video data, we obtain a fixed dimensional representation of each frame from a pretrained ResNet~\cite{he2016deep}.


\subsection{Proposed Attentive Modality Hopping Mechanism}
We propose a novel iterative attention process, referred to as an attentive modality hopping mechanism (\textbf{AMH}), that aggregates the salient information over each modality to predict the emotion of the speech.
Figure~\ref{fig_AMH} shows the architecture of the proposed \textbf{AMH} model.
Previous research used multimodal information independently using a neural network model by fusing information over each modality~\cite{yoon2018multimodal,sebastian2019fusion}.
Recently, researchers also investigated an inter-attention mechanism over the modality~\cite{xu2019learning,yoon2019speech}.
As opposed to this approach, we propose a neural network architecture that aggregates information in a modality by conditioning on other modalities via an iterative process.

First, the sequential features of each modality are encoded using the recurrent encoder by equation~(\ref{eq_RE}).
Then, the last-step hidden states of the audio recurrent encoder, $\textbf{h}^{A}_{\text{last}}$, and the text recurrent encoder, $\textbf{h}^{T}_{\text{last}}$, are fused together to form contextual knowledge, $\textbf{C}$.
We then apply an attention mechanism to the video sequence, $\textbf{h}^{V}_{\text{t}}$, to aggregate the salient part of the video modality.
As this model is developed with a single attention method, we refer to the model as \textbf{AMH-1}.
The final result of the \textbf{AMH-1} model, $\textbf{H}_{\text{hop1}}$, is calculated as follows:
\begin{equation}
\begin{aligned}
    &\textbf{H}_{\text{hop1}}=[~\textbf{h}^{A}_{\text{last}};\textbf{h}^{T}_{\text{last}};\textbf{H}_{1}^{V}~], \\
    &\textbf{H}_{1}^{V}={\sum_{i}} a_{i}~{\textbf{h}^{V}_{i}}, \\
    &a_i=\dfrac{\text{exp}({~(\textbf{C})}^\intercal\,\textbf{W}\,\textbf{h}^{V}_{i}~)}{\sum_{i} \text{exp}({~(\textbf{C})}^\intercal\,\textbf{W}\,\textbf{h}^{V}_{i}~)},~(i=1,...,t), \\
    &\textbf{C}=f(~{\textbf{h}^{A}_{\text{last}}}, {\textbf{h}^{T}_{\text{last}}}~),
\end{aligned}
\label{eq_hop_1}
\end{equation}
where $f$ is a fusion function (we use a vector-concatenation in this study), and $\textbf{W}\,{\in}\,\mathbb{R}^{d\times d}$ and bias $\textbf{b}$ are learned model parameters. The overall flow is presented in figure~\ref{fig_hop_1}.

The $\textbf{H}_{1}^{V}$ in equation~(\ref{eq_hop_1}) is a new modality representation for visual information considering the audio and text modality.
With this information, we apply the 2nd attentive modality hopping process, referred to as \textbf{AMH-2}, to the audio sequence.
The final result of the \textbf{AMH-2} model, $\textbf{H}_{\text{hop2}}$, is calculated as follows:
\begin{equation}
\begin{aligned}
  &\textbf{H}_{\text{hop2}}=[~\textbf{H}_{1}^{A};\textbf{h}^{T}_{\text{last}};\textbf{H}_{1}^{V}~], \\
  &\textbf{H}_{1}^{A}={\sum_{i}} a_{i}~{\textbf{h}^{A}_{i}}, \\
	&a_i=\dfrac{\text{exp}({~(\textbf{C})}^\intercal\,\textbf{W}\,\textbf{h}^{A}_{i}~)}{\sum_{i} \text{exp}({~(\textbf{C})}^\intercal\,\textbf{W}\,\textbf{h}^{A}_{i}~)},~(i=1,...,t), \\
	&\textbf{C}=f(~{\textbf{h}^{T}_{\text{last}}}, \textbf{H}_{1}^{V}~),
\end{aligned}
\label{eq_hop_2}
\end{equation}
where $\textbf{H}_{1}^{A}$ is a new representation for audio information considering the textual and visual modality after the \textbf{AMH-1} process.

We further apply the 3rd attentive modality hopping process to the textual sequence, referred to as \textbf{AMH-3}, with the updated audio and visual representations from the equations~(\ref{eq_hop_1}) and (\ref{eq_hop_2}).
The final result of the \textbf{AMH-3} model, $\textbf{H}_{\text{hop3}}$, is calculated as follows:
\begin{equation}
\begin{aligned}
  &\textbf{H}_{\text{hop3}}=[~\textbf{H}_{1}^{A};\textbf{H}_{1}^{T};\textbf{H}_{1}^{V}~], \\
  &\textbf{H}_{1}^{T}={\sum_{i}} a_{i}~{\textbf{h}^{T}_{i}}, \\
	&a_i=\dfrac{\text{exp}({~(\textbf{C})}^\intercal\,\textbf{W}\,\textbf{h}^{T}_{i}~)}{\sum_{i} \text{exp}({~(\textbf{C})}^\intercal\,\textbf{W}\,\textbf{h}^{T}_{i}~)},~(i=1,...,t), \\
	&\textbf{C}=f(~\textbf{H}_{1}^{A}, \textbf{H}_{1}^{V}~),
\end{aligned}
\label{eq_hop_3}
\end{equation}
where $\textbf{H}_{1}^{T}$ is an updated representation of the textual information considering the audio and visual modalities.
Similarly, we can repeat the \textbf{AMH-1} process with updated modalities, $\textbf{H}_{1}^{A}$, $\textbf{H}_{1}^{T}$ and $\textbf{H}_{1}^{V}$, to define the \textbf{AMH-4} process and compute $\textbf{H}_{\text{hop4}}$ as follows:
\begin{equation}
\begin{aligned}
    &\textbf{H}_{\text{hop4}}=[~\textbf{H}_{1}^{A};\textbf{H}_{1}^{T};\textbf{H}_{2}^{V}~], \\
    &\textbf{H}_{2}^{V}={\sum_{i}} a_{i}~{\textbf{h}^{V}_{i}}, \\
    &a_i=\dfrac{\text{exp}({~(\textbf{C})}^\intercal\,\textbf{W}\,\textbf{h}^{V}_{i}~)}{\sum_{i} \text{exp}({~(\textbf{C})}^\intercal\,\textbf{W}\,\textbf{h}^{V}_{i}~)},~(i=1,...,t), \\
    &\textbf{C}=f(~\textbf{H}_{1}^{A}, \textbf{H}_{1}^{T}~).
\end{aligned}
\label{eq_hop_4}
\end{equation}

As the proposed \textbf{AMH} mechanism employs an iterative modality hopping process, we can generalize the \textbf{AMH-N} formulation that allows the model to hop \textit{N}-times over the modality by repeating the equations~(\ref{eq_hop_1}), (\ref{eq_hop_2}), and (\ref{eq_hop_3}) in order.

\subsection{Optimization}
\label{ssec:optimization}
Because our objective is to classify speech emotion, we pass the final result of \textbf{AMH-N}, $\textbf{H}_{\text{hopN}}$, through the softmax function to predict the seven-category emotion class.
We employ the cross-entropy loss function as defined by:
\begin{equation}
\begin{aligned}
 &\hat{y}_{c} = \text{softmax}((\textbf{H}_{\text{hopN}})^\intercal~\textbf{W}+\textbf{b}~), \\
 &\mathcal{L} = {\frac{1}{N}} \sum_{j=1}^{N} \sum_{c=1}^{C} y_{j,c} \text{log} (\hat{y}_{j,c}),
\end{aligned}
\label{eq_loss}
\end{equation}
where $y_{j,c}$ is the true label vector, and $\hat{y}_{j,c}$ is the predicted probability from the softmax layer.
The $\textbf{W}$ and the bias $\textbf{b}$ are model parameters. $C$ is the total number of classes, and $N$ is the total number of samples used in training.

\section{Experiments}
\label{sec:experiments}
\subsection{Dataset and Experimental Setup}

We use the Interactive Emotional Dyadic Motion Capture (IEMOCAP)~\cite{busso2008iemocap} dataset, which contains abundant multimodal emotion descriptions of natural utterances.
The corpus includes five sessions of utterances between two speakers (one male and one female). A total of 10 unique speakers participated in this work.
The emotional category for each utterance was annotated by three people.
First, we eliminated all data that were labeled as three different emotions.
Then, following previous research~\cite{yoon2018multimodal,cho2018deep,yoon2019speech}, we assign a single categorical emotion to the utterance in which the majority of annotators agreed on the emotion labels.
The final dataset contains 7,487 utterances in total (1,103 \textit{angry}, 1,041 \textit{excite}, 595 \textit{happy}, 1,084 \textit{sad}, 1,849 \textit{frustrated}, 107 \textit{surprise} and 1,708 \textit{neutral}). We do not include classes from the original dataset that is too small in size, such as 3 \textit{other}, 40 \textit{fear} and 2 disgust.
In the training process, we perform 10-fold cross-validation where each 8-, 1-, and 1-fold are used for the training set, development set, and test set, respectively.

\subsection{Feature extraction and Implementation details}
As this research is extended work from previous research~\cite{yoon2019speech}, we use the same feature extraction method for audio and text modality as in our previous work.
After extracting the 40-dimensional mel-frequency cepstral coefficients (MFCC) feature (frame size is set to 25 ms at a rate of 10 ms with the Hamming window) using Kaldi~\cite{povey2011kaldi}, we concatenate it with its first- and second-order derivatives, making the feature dimension 120.


In preparing the textual dataset, we first use the ground-truth transcripts of the IEMOCAP dataset.
We further obtain the transcripts by using a commercial automatic speech recognition system~\cite{GoogleCloudSpeechAPI} (the performance of the ASR system is 5.53\% word error rate) for the practical use case where the ground-truth may not be available.
For both cases, we apply a word-tokenizer to the transcription and obtain sequential features of the textual modality.

For the visual dataset, we perform an additional prepossessing. The video data in the IEMOCAP are recorded with the two actors together in the video frame. We first split each video frame into two subframes so that each segment contains only one actor.
Then, we crop the center of each frame with size 224*224 to focus on the actor and to remove the background in the video frame. Finally, we extract 2,048-dimensional visual features from each subframe using pretrained ResNet-101~\cite{he2016deep} at a frame rate of 3 per second.

We minimize the cross-entropy loss function (equation~(\ref{eq_loss})) using the Adam optimizer~\cite{kingma2014adam} with a learning rate of 1e-3 and gradients clipped with a norm value of 1.
All of the code developed for the empirical results is available via web repository.\footnote{http://github.com/david-yoon/attentive-modality-hopping-for-SER}

\begin{table}[!t]
\small
\centering
\begin{tabular}
{C{0.28\columnwidth}C{0.15\columnwidth}C{0.185\columnwidth}C{0.19\columnwidth}}
\toprule
\textbf{Model} & \textbf{Modality} & \textbf{WA}  & \textbf{UA} \\ 
\midrule
\multicolumn{4}{c}{Ground-truth transcript} \\
\midrule
\textbf{RNN-attn}~\cite{mirsamadi2017automatic}	   & A	& 0.359~{\scriptsize$\pm \text{0.058}$} & 0.319~{\scriptsize$\pm \text{0.065}$}  \\
\textbf{RNN-attn}~\cite{mirsamadi2017automatic}	   & T	& \textbf{0.528}~{\scriptsize$\pm \textbf{0.018}$} & \textbf{0.525}~{\scriptsize$\pm \textbf{0.031}$}   \\
\textbf{RNN-attn}~\cite{mirsamadi2017automatic}	   & V	& 0.467~{\scriptsize$\pm \text{0.046}$} & 0.412~{\scriptsize$\pm \text{0.066}$}   \\

\midrule
\textbf{MDRE}~\cite{yoon2018multimodal} & A+T	& 0.557~{\scriptsize$\pm \text{0.018}$} & 0.536~{\scriptsize$\pm \text{0.030}$} \\
\textbf{MDRE}~\cite{yoon2018multimodal} & T+V	& 0.585~{\scriptsize$\pm \text{0.040}$} & 0.561~{\scriptsize$\pm \text{0.046}$} \\
\textbf{MDRE}~\cite{yoon2018multimodal} & A+V	& 0.481~{\scriptsize$\pm \text{0.049}$} & 0.415~{\scriptsize$\pm \text{0.047}$} \\
\textbf{MHA}~\cite{yoon2019speech} 	   & A+T	& 0.583~{\scriptsize$\pm \text{0.025}$} & 0.555~{\scriptsize$\pm \text{0.040}$} \\
\textbf{MHA}~\cite{yoon2019speech} 	   & T+V	& \textbf{0.590}~{\scriptsize$\pm \textbf{0.017}$} & \textbf{0.560}~{\scriptsize$\pm \textbf{0.032}$} \\
\textbf{MHA}~\cite{yoon2019speech} 	   & A+V	& 0.490~{\scriptsize$\pm \text{0.049}$} & 0.434~{\scriptsize$\pm \text{0.060}$} \\

\midrule
\textbf{MDRE}~\cite{yoon2018multimodal} 	   & A+T+V	& 0.602~{\scriptsize$\pm \text{0.033}$} & 0.575~{\scriptsize$\pm \text{0.046}$} \\
\textbf{AMH} (ours) 	   & A+T+V	& \textbf{0.624}~{\scriptsize$\pm \textbf{0.022}$} & \textbf{0.597}~{\scriptsize$\pm \textbf{0.040}$} \\
\midrule
\multicolumn{4}{c}{ASR-processed transcript} \\
\midrule
\textbf{AMH-\textit{ASR}} (ours) 	   & A+T+V	&
\text{0.611}~{\scriptsize$\pm \text{0.024}$} & \text{0.595}~{\scriptsize$\pm \text{0.036}$} \\
\bottomrule 
\end{tabular}
\caption{
Model performance comparisons.
The ``A'', ``T'' and  ``V'' in modality indicate ``Audio'', ``Text'' and ``Video'', receptively.
The ``-\textit{ASR}'' models use ASR-processed transcripts.
}
\label{table_performance}
\end{table}

\subsection{Performance evaluation}

We report the model performance using the weighted accuracy (WA) and unweighted accuracy (UA).
We perform 10-fold cross-validation (10 experiments for each fold) and report the average and standard deviation results.

Table~\ref{table_performance} shows the model performance on the speech emotion recognition task.
To compare our results with previous approaches, we report the model performance in regards to the types of modalities used for the experiments.
From the previous model, \textbf{RNN-attn} employs attention memory to compute and aggregate the emotionally salient part while encoding a single modality using a bidirectional LSTM network. It achieves the best performance of 0.528 WA, with the use of the textual modality.
In contrast, the \textbf{MDRE} model use multiple RNNs to encode multiple modalities and merge the results using another fully connect neural network layer. Similarly, the \textbf{MHA} model employs dual-RNN for any two modalities and compute inter-attention over each modality.
Among the models that use any two modalities, \textbf{MHA} achieves the best performance of 0.590 WA, with the textual and visual modalities.

Finally, we evaluate model performances with the use of three modalities, audio, text, and video. Our proposed model, \textbf{AMH}, outperforms \textbf{MDRE} by 3.65\% relative (0.602 to 0.624 absolute) in terms of WA.
Note that we omit the \textbf{MHA} experiments with three modalities since they cannot deal with three modalities.
In a practical scenario, we may not access the audio transcripts.
We describe the effect of using ASR-processed transcripts on the proposed system. 
As shown in table~\ref{table_performance}, we observe performance degradation in \textbf{AMH-ASR} compared to that of \textbf{AMH} (our best system) by 2.08\% (0.624 to 0.611) relative in WA.
Even with the erroneous transcripts, however, the proposed system \textbf{AMH-ASR} surpasses the baseline system (\textbf{MDRE}) by 1.49\% relative (0.602 to 0.611) in terms of WA.

\begin{table}[!t]
\small
\centering
\begin{tabular}
{C{0.10\columnwidth}C{0.15\columnwidth}C{0.10\columnwidth}C{0.20\columnwidth}C{0.20\columnwidth}}
\toprule
\textbf{\# hop} & \textbf{context} & \textbf{target} & \textbf{WA}  & \textbf{UA} \\ 

\midrule
1 & $\textbf{h}_{\text{last}}^{A}$~~$\textbf{h}_{\text{last}}^{T}$ & V &0.599~{\scriptsize$\pm \text{0.015}$} & 0.579~{\scriptsize$\pm \text{0.032}$} \\
2 & $\textbf{h}_{\text{last}}^{T}$~~$\textbf{H}_{1}^{V}$ & A &0.610~{\scriptsize$\pm \text{0.024}$} & 0.589~{\scriptsize$\pm \text{0.047}$} \\
3 & $\textbf{H}_{1}^{A}$~~$\textbf{H}_{1}^{V}$ & T &0.605~{\scriptsize$\pm \text{0.033}$} & 0.581~{\scriptsize$\pm \text{0.051}$} \\
\midrule
\textbf{4} & $\textbf{H}_{1}^{A}$~~$\textbf{H}_{2}^{T}$ & V &\textbf{0.612}~{\scriptsize$\pm \textbf{0.030}$} & \textbf{0.586}~{\scriptsize$\pm \textbf{0.041}$} \\
5 & $\textbf{H}_{2}^{T}$~~$\textbf{H}_{2}^{V}$ & A &0.600~{\scriptsize$\pm \text{0.038}$} & 0.581~{\scriptsize$\pm \text{0.054}$} \\
6 & $\textbf{H}_{2}^{A}$~~$\textbf{H}_{2}^{V}$ & T &0.610~{\scriptsize$\pm \text{0.023}$} & 0.584~{\scriptsize$\pm \text{0.046}$} \\
\midrule
\textbf{7} & $\textbf{H}_{2}^{A}$~~$\textbf{H}_{3}^{T}$ & V &\textbf{0.624}~{\scriptsize$\pm \textbf{0.022}$} & \textbf{0.597}~{\scriptsize$\pm \textbf{0.040}$} \\
8 & $\textbf{H}_{3}^{T}$~~$\textbf{H}_{3}^{V}$ & A &0.603~{\scriptsize$\pm \text{0.034}$} & 0.576~{\scriptsize$\pm \text{0.047}$} \\
9 & $\textbf{H}_{3}^{A}$~~$\textbf{H}_{3}^{V}$ & T &0.608~{\scriptsize$\pm \text{0.034}$} & 0.583~{\scriptsize$\pm \text{0.048}$} \\
\bottomrule 
\end{tabular}
\caption{
Model (\textbf{AMH}) performance as the number of hop increases. Top-2 scores marked in bold.
}
\label{table_hop_performance}
\end{table}



\begin{figure}[!t]
\small
\centering
\includegraphics[width=0.84\columnwidth]{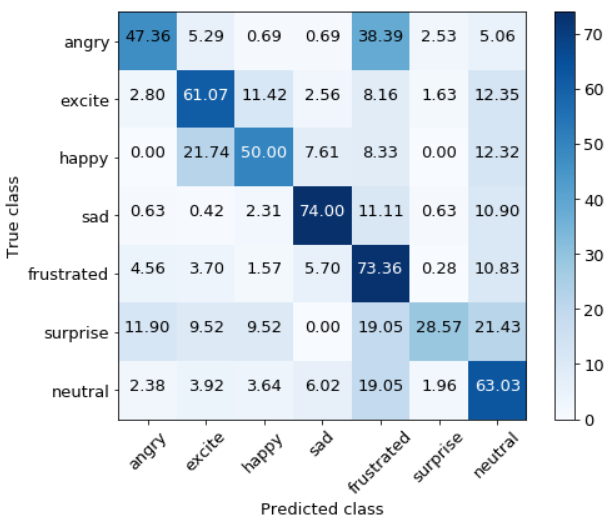}
\caption{
Confusion matrix of the AMH-7 model
}
\label{fig_confusion}
\end{figure}

Table~\ref{table_hop_performance} shows the model performance (\textbf{AMH}) as the number of hops increases. 
As the number of hops increases, the representative vector of each modality gets updated in order according to equation~(\ref{eq_hop_1})-(\ref{eq_hop_3}). 
We find that the model achieves the best performance in the 7-hop case. This behavior demonstrates that our proposed model correctly learns to aggregate the salient information among the modalities via the iterative hopping process.

\subsection{Error analysis}

Figure~\ref{fig_confusion} shows the confusion matrices of the proposed systems.
In general, most of the emotion labels are frequently misclassified as \textit{neutral} class, supporting the claims of~\cite{yoon2019speech,neumann2017attentive}.
The model confused the \textit{excite} and \textit{happy} classes since there exists a report of overlap in distinguishing these two classes, even in human evaluations~\cite{busso2008iemocap}.
It is interesting to observe that the model misclassifies the \textit{angry} class to the \textit{frustrated} class with a rate of 0.383; however, the rate is 0.045 in the opposite case (confusion \textit{frustrated} to \textit{angry} class).
It is natural to see that the model is the most inaccurate in the \textit{surprise} class because we only have a small-size dataset for that class (107 samples).

\section{Conclusions}
\label{sec:conclusions}
In this paper, we proposed an attentive modality hopping mechanism for speech emotion recognition tasks.
The proposed mechanism computes the salient part and aggregates the sequential information from one modality by conditioning on the other two modalities via an iterative hopping process.
Extensive experiments demonstrate that the proposed \textbf{AMH} surpasses the best baseline system by 3.65\% relative improvement in terms of weighted accuracy.

\section*{Acknowledgments}
K. Jung is with ASRI, Seoul National University, Korea. 
This work was supported by MOTIE, Korea, under Industrial Technology Innovation Program (No.10073144) and by the NRF grant funded by the Korea government (MSIT) (NRF2016M3C4A7952587).



\bibliographystyle{IEEEbib}
\bibliography{ICASSP20}


\end{document}